\documentclass[lettersize,journal]{IEEEtran}
\usepackage{amsmath,amsfonts}
\usepackage{algorithmic}
\usepackage{algorithm}
\usepackage{array}
\usepackage[caption=false,font=normalsize,labelfont=sf,textfont=sf]{subfig}
\usepackage{textcomp}
\usepackage{stfloats}
\usepackage{url}
\usepackage{verbatim}
\usepackage{graphicx}
\usepackage{cite}
\usepackage{xcolor}
\usepackage{mathrsfs}
\usepackage{graphicx}
\usepackage{amssymb}
\newcommand{\argmin}{\arg\!\min}
\newcommand{\Amat}{{\bf A}}
\newcommand{\Bmat}{{\bf B}}
\newcommand{\Cmat}{{\bf C}}

\newcommand{\Hmat}[0]{{{\bf H}}}
\newcommand{\Imat}{{\bf I}}

\newcommand{\Kmat}[0]{{{\bf K}}}

\newcommand{\Mmat}[0]{{{\bf M}}}

\newcommand{\Qmat}[0]{{{\bf Q}}}

\newcommand{\Tmat}[0]{{{\bf T}}}

\newcommand{\Wmat}[0]{{{\bf W}}}
\newcommand{\Xmat}{{\bf X}}

\newcommand{\Zmat}{{\bf Z}}

\newcommand{\av}{\boldsymbol{a}}

\newcommand{\bv}{\boldsymbol{b}}

\newcommand{\hv}[0]{{\boldsymbol{h}}}

\newcommand{\mv}[0]{{\boldsymbol{m}}}

\newcommand{\sv}[0]{{\boldsymbol{s}}}

\newcommand{\xv}{\boldsymbol{x}}

\newcommand{\zv}{\boldsymbol{z}}

\newcommand{\Phimat}{\boldsymbol{\Phi}}

\newcommand{\betav}[0]{{\boldsymbol{\beta}} }
\newcommand{\gammav}{{\boldsymbol{\gamma}}}

\newcommand{\thetav}{\boldsymbol{\theta}}

\newcommand{\piv}{\boldsymbol{\pi}}

\usepackage{colortbl}
\definecolor{mygray}{gray}{.88}
\usepackage{amsthm} 

\usepackage{tablefootnote}
\usepackage{booktabs}
\usepackage{hyperref}
\usepackage{amsmath}
\usepackage{threeparttable}
\usepackage{wrapfig}
\usepackage{amssymb}
\usepackage{multirow}
\usepackage{graphicx}
\usepackage[section]{placeins}
\usepackage{booktabs}
\usepackage{arydshln}
\usepackage{algorithm}
\usepackage{algorithmic}

\begin{document}

\title{Investigating Mask-aware Prototype Learning for Tabular Anomaly Detection}

\author{Ruiying Lu, Jinhan Liu, Chuan Du, Dandan Guo*,~\IEEEmembership{Member,~IEEE,}
\thanks{\IEEEcompsocthanksitem This work was supported by the National Natural Science Foundation of China (NSFC) under Grant 62306125, the Natural Science Basic Research Plan in Shaanxi Province of China under Grant [2024JC-YBQN-0661],  and the Nanning Scientific Research and Technological Development Project (20231042).
\IEEEcompsocthanksitem Corresponding author*: Dandan Guo.
\IEEEcompsocthanksitem Ruiying Lu is with the School of Cyber Engineering, Xidian University, Xi’an 710071, Shaanxi, P. R. China, (e-mail: luruiying@xidian.edu.cn).
\IEEEcompsocthanksitem Chuan Du is with the School of Electronic and Information Engineering, Nanjing University of Information Science and Technology, Nan jing 210044, China (E-mail:  hnlyjzdc@163.com).
\IEEEcompsocthanksitem  Jinhan Liu and Dandan Guo are with the School of Artificial Intelligence, Jilin University, Jilin. (E-mail: jinhan23@mails.jlu.edu.cn; guodandan@jlu.edu.cn.) }}



\maketitle

\begin{abstract}
Tabular anomaly detection, which aims at identifying deviant samples, has been crucial in a variety of real-world applications, such as medical disease identification, financial fraud detection, intrusion monitoring, etc. Although recent deep learning-based methods have achieved competitive performances, these methods suffer from representation entanglement and the lack of global correlation modeling, which hinders anomaly detection performance. To tackle the problem, we incorporate mask modeling and prototype learning into tabular anomaly detection. The core idea is to design learnable masks by disentangled representation learning within a projection space and extracting normal dependencies as explicit global prototypes. Specifically, the overall model involves two parts: (i) During encoding, we perform mask modeling in both the data space and projection space with orthogonal basis vectors for learning shared disentangled normal patterns; (ii) During decoding, we decode multiple masked representations in parallel for reconstruction and learn association prototypes to extract normal characteristic correlations. Our proposal derives from a distribution-matching perspective, where both projection space learning and association prototype learning are formulated as optimal transport problems, and the calibration distances are utilized to refine the anomaly scores. Quantitative and qualitative experiments on 20 tabular benchmarks demonstrate the effectiveness and interpretability of our model.
\end{abstract}


\section{Introduction}

Tabular data, often structured as tables in relational databases with rows signifying individual data samples and columns representing feature variables, have become indispensable across diverse real-world domains including intrusion detection in cybersecurity~\cite{tifs9319404, ahmad2021network}, engineering~\cite{ye2023uadb}, finance~\cite{assefa2020generating} etc. Tabular anomaly detection (AD), which endeavors to identify samples that diverge from a pre-defined notion of normality, playing a pivotal role in diverse scientific and industrial contexts, such as medical disease identification~\cite{fernando2021deep}, financial fraud detection~\cite{al2021financial}, cybersecurity intrusion monitoring~\cite{tifs9865986, malaiya2019empirical}, and astronomy~\cite{reyes2020transformation}.
In practical scenarios, obtaining labeled anomalies is usually impractical or prohibitive, necessitating a common implementation of training solely on normal samples, called the `one-class' setting. Under this circumstance, by distilling the inherent characteristic patterns from normal training data, anomalies are expected to be detected with deviations from normal patterns~\cite{ruff2021unifying,Tifs9926133,Tifs8649753}. Nevertheless, the intricate, heterogeneous, and unstructured nature of tabular data features~\cite{chang2023data} poses significant challenges in identifying such characteristic patterns.

Recent works~\cite{qiu2021neural, shenkar2022anomaly, thimonier2024beyond} have highlighted the importance of considering the particular characteristics of tabular data. For example, Neutral AD~\cite{qiu2021neural} and {ICL}~\cite{shenkar2022anomaly} employ contrastive learning-based loss functions to create pretext tasks for tabular data, where the characteristic patterns are modeled by the contrastive losses and samples with a high loss value indicate a high possibility of anomaly. Recently, several models adhere to the reconstruction pipeline to capture characteristic patterns during reconstruction, which achieves state-of-the-art (SOTA) performances for tabular anomaly detection. In particular, NPT-AD~\cite{thimonier2024beyond} leverages Non-Parametric Transformers (NPT)~\cite{kossen2021self} to capture both feature-feature and sample-sample dependencies for anomaly detection during reconstructing tabular data. MCM~\cite{yin2024mcm} designs a mask cell modeling method to capture intrinsic correlations between features in training data and detect anomalies with reconstruction errors. 
{Typically, the motivation behind these methodologies is that a well-trained model struggles to generate or represent samples that deviate significantly from the normal distribution~\cite{yin2024mcm,pang2021deep}.} 
To be more specific, reconstruction-based methods detect anomalies by assuming that a well-trained model (only trained with normal samples) always produces normal samples regardless of whether the anomalies are within the inputs. Under this assumption, there will be larger reconstruction errors for anomalous samples, making them distinguishable from the normal ones with lower reconstruction errors. 

Despite achieving some commendable performances, when encountering complex data distribution, such as heterogeneous unstructured tabular data, reconstruction methods tend to confuse normal and anomaly data with similar reconstruction errors, resulting in the failure of anomaly detection. 
The inherent reasons might reside in the representation entanglement, where different features or relations of the learned data representations are highly correlated and entangled with each other, impeding anomaly discriminability and precise anomaly detection. Furthermore, the representations tend to overlook global correlation patterns as each data sample is represented distinctively, which fails to model the shared normal information among distinct normal samples and thus hinders the detection performance~\cite{ye2024ptarl}.


To tackle the above issues, we introduce PTAD, a prototype-oriented tabular anomaly detection method for tabular AD. The fundamental concepts center on two key aspects: 1) To learn disentangled representations and global shared normal patterns, we propose multiple masking strategies with disentangled prototype learning. 
Specifically, a data-space soft masking strategy and a projection-space multiple masking strategy are developed to select optimal masks, which facilitate the capture of various data characteristics and diverse inherent relationships. Furthermore, to encourage disentangled representation learning, projection space is constructed based on a group of learnable orthogonal basis vectors. 
2) Considering the characteristics of tabular data are heterogeneous and complex, we investigate the correlation patterns between features to facilitate modeling of tabular normal patterns and detecting anomalies, termed association prototype learning. The processes of basis vectors learning and association prototypes learning are formulated as optimal transport (OT) problems from a distribution-matching perspective, in which the transport cost can naturally serve as a criterion for anomaly assessment as it detects the deviation degree from the learned normality patterns.



In brief, our main contributions are summarized as follows: 

(1) Considering the intricate, heterogeneous, and unstructured nature of tabular data, we develop a multiple mask modeling method to capture intrinsic correlations, derived in both data space and projection space.

{(2) We encourage learning the disentangled representation within the projection space and introduce a group of learnable orthogonal basis vectors for extracting normal feature-level patterns.}

{(3) For compensation for feature-level patterns, we further investigate the correlation patterns across features by introducing a novel prototype association learning method to facilitate modeling normal cross-feature patterns and detecting anomalies.}


{(4) Extensive quantitative and qualitative experiments on various datasets demonstrate the superiority and effectiveness of our method for tabular AD.}

\section{Related Work}

\textbf{Tabular Anomaly Detection.}
Over the past decades, numerous methods for tabular AD have been developed to identify significant deviations from the majority of data objects, which can be roughly divided into four groups:  
\textbf{\textit{i) Supervised methods.}} With the availability of both normal and abnormal training samples, supervised methods such as Support Vector Machine (SVM)~\cite{hearst1998support} and deep networks~\cite{gorishniy2021revisiting} developed, however, facing the risk of missing unknown anomalies. 
\textbf{\textit{ii) Semi-supervised methods.}} Capitalizing the supervision from partial labels, the semi-supervised algorithms~\cite{pang2023deep,Tifs9926133} efficiently use the partially labeled data and facilitate representation learning with the unlabeled data. 
\textbf{\textit{iii) Unsupervised methods.}} Without any label information of training data, unsupervised methods aim to find deviations from the majority of data, e.g. deep autoencoders~\cite{kim2019rapp,han2022adbench} and GANs~\cite{schlegl2017unsupervised,sabuhi2021applications} suppose abnormality can be indicated with high reconstruction error. 
\textbf{\textit{iv) Self-Supervised method.}}
Several recent studies have revealed that self-supervised learning facilitates anomaly detection by creating pretext tasks to train neural networks for modeling better characteristics within training data. 
{In particular, NPT-AD~\cite{thimonier2024beyond} leverages Non-Parametric Transformers for anomaly detection to capture both feature-feature and sample-sample dependencies while reconstructing tabular data. Additionally, MCM~\cite{yin2024mcm} extends the mask modeling to tabular AD, which generates diverse multiple masks and jointly utilizes its reconstructions for anomaly detection.} 
However, reconstruction models usually suffer from representation entanglement and limited global correlation modeling, and the reconstruction error as a general anomaly detection score, is limited for clear and precise anomaly detection. This motivates us to perform mask modeling and prototype learning to relieve anomaly reconstruction and find a new indicator for anomaly scoring.


\textbf{Prototype Learning.} Prototype learning has been widely studied in different tasks of computer vision~\cite{nauta2021neural,zhou2022rethinking}, and natural language processing~\cite{zalmout2022prototype}.
Typically, prototypes refer to empirical proxies and are computed as the weighted results of latent features of all instances of a particular class, and the distances to prototypes facilitate classification, recognition, representations, etc.
Recently, prototype learning has been introduced to image anomaly detection to facilitate extracting normal feature representations to distinguish anomalous samples. In particular, HVQ-Trans~\cite{lu2023hierarchical} preserves the typical normal features as discrete iconic prototypes for image reconstruction via vector quantization.
Furthermore, VPDM~\cite{livague} leverages prototypes as vague information about the target into a conditional diffusion model to incrementally enhance details for reconstruction.
However, tabular data exhibits heterogeneous, intricate features devoid of a rigid structure~\cite{chang2023data}, posing significant challenges in identifying distinctive characteristic patterns. Simply adopting a straightforward approach to extract feature prototypes is inadequate for tabular data. Consequently, we are motivated to learn the intricate correlation patterns among features, termed association prototypes, rather than focusing solely on the features themselves, to enhance the capabilities of tabular AD.




\section{Preliminary}

\textbf{Problem Formulation.} This paper aims at tabular AD, where the training set only contains normal samples following the one-class classification setting. Denoting the training set of \( N_{train} \) in-class normal samples as \( D_{\text{train}} = \{ \mathbf{x}_n \}_{n=1}^{N_{train}} \), where each sample is a \( d \)-dimensional vector. Denoting the testing set of \( N_{test} \) samples as \( D_{\text{test}} = \{ \mathbf{x}_n \}_{n=1}^{N_{test}}\), which contains both normal and abnormal samples. The objective of tabular AD is to develop an anomaly scoring function \( \mathcal{S}: \mathbb{R}^d \to \mathbb{R} \) that assigns low scores to samples drawn from the same underlying distribution as \( D_{\text{train}} \) and high scores to the samples not aligned with  \( D_{\text{train}} \). Typically, standard reconstruction-based approaches~\cite{yin2024mcm,thimonier2024beyond} learn a mapping function $\Phimat_\theta: \mathbb{R}^d \longrightarrow \mathbb{R}^d$ by minimizing the reconstruction loss, which is often employed as the measurement of anomaly score.

\textbf{Non-Parametric Transformer.} NPT has shown the priority of reasoning about relationships between both datapoints and features~\cite{kossen2021self,thimonier2024beyond} for tabular data. NPT receives the data $\Xmat \in \mathbb{R}^{N\times d }$ and stochastic masking matrix with the same dimention as input, then maps them through a linear mapping into $\Hmat^0 \in \mathbb{R}^{N\times d \times e}$ by transforming each feature of each sample in data space into an $e$-dimensional embedding. Each NPT layer involves an attention between datapoints (ABD) layer and an attention between attributes (ABA) layer to capture sample-sample and feature-feature dependencies, respectively, formulated as follows: 
\begin{equation}
\centering
\begin{aligned}
    \mbox{ABD}(\Hmat^l) &= \mbox{MHSA}(\Hmat^l) = \Hmat^{l+1} \in \mathbb{R}^{{N\times H}},\\
    \mbox{ABA}(\Hmat^l) &= \underset{\text {{axis=$N$}}}{\text{Stack}} \left(\mbox{MHSA}(\Hmat^l_1), . . . , \mbox{MHSA}(\Hmat^l_{{N}})\right) \\
    & = \Hmat^{l+1} \in \mathbb{R}^{{N\times H}},
\end{aligned}
\end{equation}
where $l$ denotes the layer index. For the ABD layer, the embedding can be flattened to $\mathbb{R}^{N\times H}$ with $H=d\times e$ and the multi-head self-attention (MHSA) is applied across all samples. For the ABA layer, we reshape the embedding as $\mathbb{R}^{N\times d \times e}$ and then apply MHSA independently to each row (i.e. a single datapoint) across the feature dimension. By alternatively conducting ABD and ABA, NPT is trained to reconstruct the stochastic masked input and model intrinsic dependencies among datapoints and within each datapoint. 
{Motivated by NPT~\cite{kossen2021self}, \cite{thimonier2024beyond} introduce NPT-AD by incorporating both sample-sample and feature-feature dependencies in tabular AD, which showcases its effectiveness and superiority for tabular data. However, it needs to combine the validation samples and the entire training set for detecting samples during inference, which results in large computation costs and potentially compromising applicability to big datasets.}

\textbf{Optimal Transport.} OT is a powerful tool for computing the distance between two distributions by measuring their minimal transport cost, which has a rich theoretical foundation~\cite{dvurechensky2018computational,courty2016optimal},. Recently, OT has drawn widespread attention in different fields, such as generative model \cite{gen1,gen2} and domain adaptation \cite{domain1,domain2}, long-tailed classification \cite{guo2022_reweighting,gao2023OTmix,gao2024distribution}, prompt learning \cite{ren2024modality}, representation learning \cite{ye2024ptarl,guo2022adaptive,guo2022set,wang2022wete}. Here, we only focus our discussion on OT for discrete probability distributions and please refer to~\cite{peyre2019computational} for more details.
Denote two discrete probability distributions over an
arbitrary space $S \in \mathbb{R}^d$ as $p =
\sum^n_{i=1} a_i \delta_{x_i}$
and $q = \sum^m_{j=1} b_j \delta_{y_j}$, where both $\av \in
\sum^n$
and $\bv \in
\sum^m$ are discrete probabilities summing to 1. The OT distance
between $p$ and $q$ is defined as
\begin{equation}
    \mbox{OT}(p,q) = \min_{\Tmat \in \Pi(p,q)} \langle \Tmat, \Cmat \rangle,
\end{equation}
where $\langle \cdot, \cdot\rangle$ is the Frobenius dot-product and $\Cmat \in \mathbb{R}^{n \times m}_{\geq 0}$
is the transport cost matrix where $C_{ij} = \mbox{Distance}(x_i, y_j )$ reflects the cost between $x_i$ and $y_j$. The transport probability matrix $\Tmat \in \mathbb{R}^{n\times m}_{\geq 0}$ is subject to $\Pi(p,q) := \{\Tmat | \sum^n_{i=1} T_{ij}=b_j, \sum^m_{j=1} T_{ij}=a_i\}$.
Above optimization often entails substantial computational expenses, and the entropic regularization $H = -\sum_{ij} T_{ij} \ln T_{ij}$ is included to reduce the computational cost while maintaining sufficient smoothness~\cite{cuturi2013sinkhorn}.

\section{Our Proposed Method}

\begin{figure*}[t!]
\centering
\includegraphics[width=1.0\textwidth]{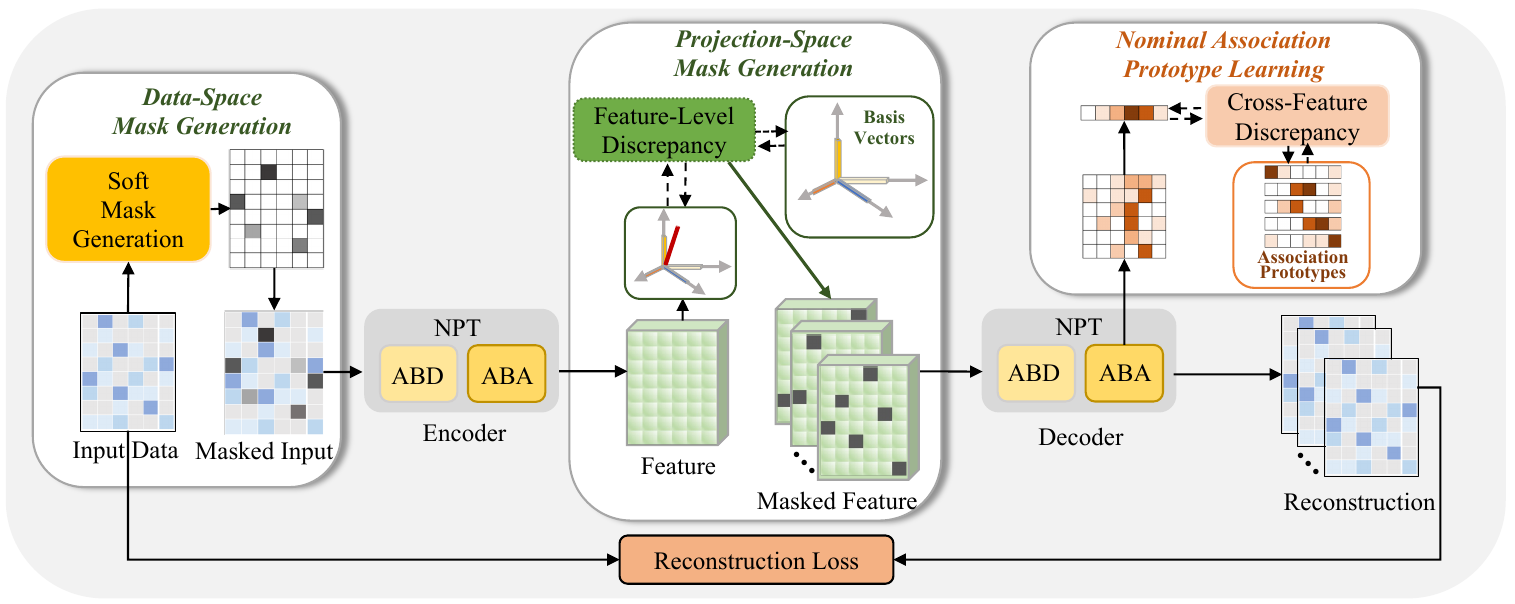}
\caption{{Overall framework: The input data is first masked with the generated soft mask in the data space and then encoded through an NPT layer. Next, in the projection space, data-adaptive multiple masks are generated for various input data and learn the disentangled feature-level discrepancies with orthogonal P-space basis vectors. Then, multiple masked features are decoded by an NPT layer, during which the association prototypes are learned to measure the cross-feature discrepancies by solving the optimal transport problem. Finally, the reconstruction errors, feature-level discrepancies, and cross-feature discrepancies are fused together to instruct anomalies.
}}
\label{fig:model}
\end{figure*}

This work follows a reconstruction pipeline, as shown in Fig.~\ref{fig:model}, which introduces data-adaptive mask modeling during encoding and association prototypes during decoding for tabular AD. Given input samples, we first generate a data-space mask and embed the masked samples with an encoder. Then, we adaptively produce various masks in the projection space according to the discrepancy between features and orthogonal basis vectors. Afterward, the decoder maps multiple masked representations from latent space to data space for reconstruction, among which we learn the normal association prototypes by aligning its distribution to the distribution over shared correlation patterns. Both the discrepancy in projection space and the alignment in decoding stage are formulated as OT problems and integrated with reconstruction loss for optimization and anomaly scoring.

\subsection{Masking Strategy}
Inspired by mask modeling in CV and NLP, we aim to incorporate masks for tabular data to capture intrinsic correlations between features, which facilitates modeling the normal characteristic patterns. However, it is challenging to manually discover such optimal masks. In the following, we introduce the learnable masking strategy both within the raw data space and the projected feature space. 

%


\textbf{Data-Space Mask Generalization.} To capture intrinsic correlations existing in the original data space of training data and eliminate redundant information, we produce a learnable soft mask for input data. Given $N$ input samples $\Xmat = \{\xv_n\}_{n=1}^N \in \mathbb{R}^{N\times d}$, the data-adaptive masking can be implemented as
\begin{equation}
\label{eq3}
   {\hat{\Xmat}} = \Xmat \odot \Mmat^{ds}, \;\;\;\;\;\; \Mmat^{ds} = \frac{1}{1+e^{-\Wmat_3(\text{Relu}(\Wmat_2(\text{Relu}(\Wmat_1 \Xmat^T))))}},
\end{equation}
where $\Wmat_1 \in \mathbb{R} {^{d\times d}}, \Wmat_2 \in \mathbb{R} {^{d\times d}}, \Wmat_3 \in \mathbb{R} {^{d\times d}}$ represent linear projections and $\odot$ refers to the element-wise multiplication. Each value of {mask matrix} $\Mmat^{ds} {\in \mathbb{R} ^{N\times d}}$ is a flexible weight between zero and one, where each row corresponds to the masking degree across different features, and each column represents the masking degree across input samples for a specific feature. 
This motivates the model to uncover the statistical correlations between masked and unmasked positions across both datapoints and features.
However, the data or feature correlations in tabular data space are often highly tangled and lack statistical global structure information, thus we need to further find another disentangled space for mask modeling.

\textbf{Projection-Space Mask Generation.} To stimulate global data correlation learning, we subsequently encode the {masked input $\hat{\Xmat} $ into a disentangled Projection Space (P-space) with an NPT layer composed of an ABD and an ABA layer, denoted as $\Zmat = \Phimat_E(\hat{\Xmat}; \thetav_E) \in \mathbb{R}^{N \times H}$, {where $\Phimat_E$ is the encoder parameterized by $\thetav_E$, and $\Zmat =\{\zv_n\}_{n=1}^{N}$ where $\zv_n \in \mathbb{R}^{H}$ is the representation of $n$-th sample.}
Intuitively, we only possess the normal tabular samples during training and we assume that they share some global intrinsic characteristic patterns in the P-space. These shared normal patterns are learnable and serve as basis vectors of the P-space, denoted by 
\begin{equation}\mathcal{B} \!=\! \{\betav^1,...,\betav^k,...,\betav^{K}\} \in \mathbb{R}^{K \times H},
\end{equation}
where $K$ is the number of basis vectors and $\betav^{k} \in \mathbb{R}^{H}$ denotes the $k^{th}$ basis vector. Typically, normal samples are close to the shared basis vectors, whereas abnormal samples are distinguished by large deviations from these vectors. Thus, we introduce $\Mmat \in \mathbb{R}^{N \times H \times K}$ to mask the suspicious anomaly information in P-space:} 
\begin{equation}
\label{eq4}
\Hmat^k = \Zmat  \odot \Mmat^{k}, \;\;\;   M_{nh}^{k} = \left\{
\begin{aligned}
1, \;\;\; & (z_{nh} - \beta^k_h)^2  \leq \mu_{n}^k, \\
0, \;\;\; & (z_{nh} - \beta^k_h)^2 \textgreater \; \mu_{n}^k,
\end{aligned}
\right.
\end{equation}
{where $ \Hmat^k\in \mathbb{R}^{N \times H}$ denotes the masked representation with $k={1:K}$, $\mu_n^k = \frac{1}{H} \sum_{h=1}^H (z_{nh} - \beta^k_h)^2 $ means a data-related threshold computed by the statistic average along the feature dimension, and mask value $M_{nh}^{k} $ is element-wisely computed by the Euclidean distance between the basis vector $\betav^{k}$ and latent representation $\zv_{n}$ at the $h$-th feature.} 
The model is motivated to restore the masked features solely relying on the unmasked normal features. Compared to straightforward random masking which may leak a large amount of abnormal information, our data-adaptive masking strategy aims to learn the optimal masks, like suspicious anomaly locations during inference. 
Intuitively, the positions with larger distances to basis vectors are considered with larger probability as anomalies. By masking these positions, the model is motivated to embed these positions with unmasked normal information, leading to normal reconstructions and large deviations for indicating anomalies.


{Furthermore, the multiple masking strategy encourages the model to reconstruct samples with various masks. Therefore, anomalies are prone to be detected by a comprehensive measurement. In contrast to masking in the original space, it is more disentangled to act within this P-Space consisting of explicitly defined basis vectors. Furthermore, this designation also saves computational consumption as the multiple setting is only needed for the subsequent decoder.}

\textbf{Projection Space learning.} In the P-space, we aim to find a group of basis vectors  $\mathcal{B}$  to capture the normal characteristics of the training data. We mathematically represent the $K$ basis vectors as a $K$-dimensional empirical uniform distribution 
\begin{equation}
Q(\mathcal{B}) = \frac{1}{K}\sum_{k=1}^K \delta_{\betav^k},
\end{equation}
where $\delta_{\betav^k}$ is Dirac function of $k^{th}$ basis vectors of the discrete distribution. Besides, we view the P-space representations of $N$ data samples 
within the training set as another $N$-dimensional discrete distribution 
\begin{equation}
P(\Zmat) = \frac{1}{N}\sum_{n=1}^N \delta_{\zv_n}. 
\end{equation}
Since $\mathcal{B}$ is viewed as the global shared characteristics of the training normal data, we can enforce the distribution $Q(\mathcal{B})$ to approximate the distribution $P(\Zmat)$ to learn the encoder and basis vectors, where we solve the projection space learning problem via distribution matching. Specifically, we first learn the transport plan by minimizing the regularized distance $ \mbox{OT}(P(\Zmat), Q(\mathcal{B}))$ between these two distributions and we design the optimization loss based on the resultant transport plan, stated as
\begin{equation}
\begin{aligned}
\label{eq5}
&\mathop{\min}\limits_{\thetav_E, \mathcal{B}}\mathcal{L}_{bv} = \sum_{n=1}^{N} \min_{k\in K} T_{nk}^{\star} C_{nk}, \\
&\text{subject to} \quad \Tmat^{\star} =\argmin _{\Tmat \in \Pi(P(\Zmat),Q(\mathcal{B}))} \langle \Tmat, \Cmat \rangle-\lambda H(\Tmat),
\end{aligned}
\end{equation}
where $H(\Tmat)$ denotes the regularized entropy in \cite{cuturi2013sinkhorn}, $\lambda > 0$ is the hyper-parameter for the entropy,
$\Cmat$ is the transport cost matrix defined as $C_{nk}=\sqrt{(\zv_n - \betav^k)^2}$, and $\Tmat$ is the transport probability matrix satisfying $\Pi(P(\Zmat),Q(\mathcal{B})) := \{ \Tmat \in \mathbb{R}^{N\times K} | \sum_{k=1}^K T_{nk} = \frac{1}{N}, \sum_{n=1}^N T_{nk} = \frac{1}{K}\}$. Notably, during training, we minimize the transport distance from the P-space representation of each sample to its corresponding nearest basis vector based on the learned transport plan. {The intuition behind this is normal representations tend to approach specific one of the global basis vectors rather than its fusions, alleviating the potential collapse that anomalous projection representations also exhibit similarity with the fusion version of basis vectors.} Since all the training data are normal and thus the learned basis vectors reflect the normal patterns, at the inference stage, anomalies are prone to deviate from the nominal distributions and thus can be detected by a larger distance even with its corresponding nearest basis vector.
Furthermore, the basis vectors in P-space are expected to maintain orthogonal independence from each other, 
which can be achieved with the soft orthogonality constraint under the standard Frobenius norm, formulated as 
\begin{equation}\label{eq6} 
\begin{aligned}
&\mathop{\min}\limits_{\mathcal{B}} \mathcal{L}_{orth} = \min || \Bmat \Bmat^T - \Imat||^2_F, \; 
\end{aligned}
\end{equation}
{where $\Bmat=\{\bv_k\}_{k=1}^{K} \in \mathbb{R}^{K \times H} $ and $\bv_k =
\frac{\betav^k}{||\betav^k||}\in \mathbb{R}^{H}$ represents the normalized basis vectors. }

\subsection{Nominal Association Prototype Learning}


{In this section, we find that learning the typical correlation patterns across features, named association prototypes, facilitates modeling normal characteristic patterns of tabular data. Typically, during training with normal data, we learn the nominal association prototypes from the attention matrix of transformers. During inference, the abnormal associations are different from the nominal association prototypes, which encourages us to incorporate a calibration distance to indicate anomalies. To this end, we learn the association prototypes and the calibration distance by solving a transport distribution matching problem.}



\textbf{Association Prototypes Learning.}  Given the $K$ masked representations $\{\Hmat^k\}_{k=1}^K$ as discussed above, we use the shared decoder to output the corresponding $\{\Xmat^{rec}_k\}_{k=1}^K$ in the data space, denoted as $\Xmat^{rec}_k = \Phimat_D(\Hmat^k;\thetav_D)$, {where the decoder $\Phimat_D$ is an NPT layer composed of an ABD and an ABA layer and parameterized by $\thetav_D$.} To learn the correlation patterns between features {in sample $\xv_n$}, we investigate its query $\Qmat \in \mathbb{R}^{d\times h_k}$ and key $\Kmat \in \mathbb{R}^{d\times h_k}$ matrices for computing attribute attention of $n$-th sample in the ABA layer, where $h_k$ is the latent dimension and $d$ is the feature number of each sample. Note that the query and key matrices can be used to compute the attention matrix in MHSA, denoted as $\Amat = \text{softmax}(\frac{\Qmat \Kmat^T}{\sqrt{h_k}})\in \mathbb{R}^{d\times d}$, which provides a comprehensive understanding of the cross-feature association in the $n$-th sample. Here, we establish the lightweight association vector of data $\xv_n$ to outline the correlations, stated as $\piv_n =\{\pi_n^1,...,\pi_n^d\} \in \mathbb{R}^d$ with $
\pi_n^i= \sum_{j=1}^{h_k} \frac{q_{ij}\cdot k_{ij}}{\sqrt{h_k}}$. Accordingly, we formulate a discrete uniform distribution of all association vectors as $P(\piv) = \sum_{n=1}^{N} \frac{1}{N} \delta_{\piv_n}$. 
Besides, we denote $M$ to-be-learned association prototypes as $\Upsilon = \{ \gammav^1,...,\gammav^M\}_{m=1}^M \in \mathbb{R}^{M \times d}$ to extract shared correlation patterns, which form another discrete uniform distribution $Q(\Upsilon) = \sum_{m=1}^{M} \frac{1}{M} \delta_{\gamma^m}$.
Similar to the P-space, we here also enforce the matching between the distribution $P(\piv)$ over the association vectors and the distribution $Q(\Upsilon)$ over association prototypes. To this end, we design an OT-based optimization objective by minimizing the transport distance from the association vector to its corresponding nearest association prototype, formulated as
\begin{equation}
\begin{aligned}
\label{eq7}
&\mathop{\min}\limits_{\thetav_D, \Upsilon}\mathcal{L}_{ap} = 
\sum_{n=1}^{N} \min_{m\in M} \hat{T}_{nm}^{\star} \hat{C}_{nm},\\
& \text{subject to} \quad \hat{\Tmat}^{\star} = 
\argmin _{\hat{\Tmat} \in \Pi(P(\piv), Q(\Upsilon))} \langle \hat{\Tmat}, \hat{\Cmat} \rangle-\lambda H(\hat{\Tmat}),
\end{aligned}
\end{equation}
where cost matrix $\hat{\Cmat} \in \mathbb{R}^{N\times M}$ is calculated by Euclidean distance, and the transport probability matrix $\hat{\Tmat} \in \mathbb{R}^{N\times M}$ satisfies:
\begin{equation}
\Pi(P(\piv), Q(\Upsilon)):= \{ \hat{\Tmat}| \sum_{n=1}^N \hat{T}_{nm}=\frac{1}{M}, \sum_{m=1}^M \hat{T}_{nm} = \frac{1}{N}\}.
\end{equation}

\textbf{Loss Function.} 
Averaging reconstruction loss of multiple masked branches of the entire samples is employed as our mean objective, stated as $\mathcal{L}_{rec} = \frac{1}{K} \sum_{k=1}^K ||\Xmat - \Xmat_k^{rec}||^2_2$. By weighted averaging over the reconstruction loss, the basis vector learning loss, the association prototype learning loss, and the P-space orthogonality constrain, the overall loss of our model can be listed as follows:
\begin{equation}
\begin{aligned}
\mathop{\min} \mathcal{L} = & \mathop{\min}\limits_{\thetav_E, \thetav_D, \Wmat_{\{1,2,3\}}} \mathcal{L}_{rec} + \lambda_{bv} \cdot \mathop{\min}\limits_{\mathcal{B}} \mathcal{L}_{bv} \\
&+ \lambda_{ap} \cdot \mathop{\min}\limits_{\Upsilon} \mathcal{L}_{ap} +\lambda_{orth} \cdot \mathop{\min}\limits_{\mathcal{B}} 
 \mathcal{L}_{orth},
\end{aligned}
\end{equation}
where $\lambda_{*}$ denote coefficients of various losses respectively, and the model is jointly trained with $\mathcal{L}$.
Overall, the algorithm workflow of our PTAD is listed in Algorithm~\ref{algorithm}. Furthermore, the training pipeline of PTAD is detailed in Appendix~\ref{sec:Training Pipeline}.

\textbf{Anomaly Scoring.} 
During inference, the reconstruction loss, typically computed as the point-wise L2 norm, is widely employed as a criterion for anomaly detection. The intuition is that the reconstruction error tends to be higher for anomalous inputs, as the model is solely trained on normal data. In our model, with the $K$ reconstructions corresponding to the multiple P-space masks, we design a more robust and comprehensive reconstruction loss by $\sv^{rec}_n = \frac{1}{K}\sum_{k=1}^K \|\xv_n-\xv^{rec}_{n,k}\|_{2}^2$. However, relying solely on the reconstruction loss can be suboptimal. This motivates us to propose a new criterion to enhance the discriminability between normal and abnormal samples.
In our model, the P-space representation dissimilarity to basis vectors indicates abnormal characteristics, and the association vector dissimilarity to normal association prototypes shows abnormal dependencies. Thus, we refine the anomaly score with the calibration costs $\sv^{ap}_n$ and $\sv^{bv}_n$, stated as:
\begin{equation}
\begin{aligned}
\label{eq8}
\sv^{cab}_n=\sv^{rec}_n + \kappa \sv^{bv}_n + \alpha \sv^{ap}_n, \\
\sv^{ap}_n=\min_{m\in M} \hat{T}^{\star}_{nm} \hat{C}_{nm}, \\
\sv^{bv}_n=\min_{k\in K} T^{\star}_{nk} C_{nk},
\end{aligned}
\end{equation}
where $\hat{T}^{\star}_{nm}$ and $T^{\star}_{nk}$ subject to Eq.~\ref{eq5} and Eq.~\ref{eq7}, respectively, and $\kappa$ and $\alpha$ are weighted coefficients.

\begin{algorithm}[H]
\caption{The Algorithm Workflow of our proposed PTAD.}
\label{algorithm}
\textbf{Input:} Training dataset $D_{train}$; \\
\textbf{Parameters:} $\thetav_E$ of the NPT Encoder $\Phimat_E$, $\thetav_D$ of the NPT Decoder $\Phimat_D$, basis vectors $\mathcal{B} = \{\betav^1,...,\betav^{K}\}_{k=1}^K $, association prototypes $\Upsilon = \{ \gammav^1,...,\gammav^M\}_{m=1}^M$, $\{\Wmat_1,\Wmat_2,\Wmat_3\}$ of data-space mask generator; \\
\textbf{Output:} Reconstructions of input tabular data;
\begin{algorithmic}[1]  
    \STATE Initialize model parameters, $\mathcal{B}$ and $\Upsilon$ randomly
    \FOR{ epoch $1,2,..., T$ }
        \STATE Sample batch of  $\Xmat \in \mathbb{R}^{N\times d}$ from input datasets $D_{train}$
        \STATE Generate Data-space masked data $\hat{\Xmat}$ by Eq.~\ref{eq3}
        \STATE Encode the $\hat{\Xmat}$ through $\Phimat_E(\hat{\Xmat}; \thetav_E)$ as the P-space representation $\Zmat$  
        \STATE Build distributions for the P-space representations and basis vectors as $P(\Zmat) = \frac{1}{N}\sum_{n=1}^N \delta_{\zv_n}$ and $Q(\mathcal{B}) = \frac{1}{K}\sum_{k=1}^K \delta_{\betav^k}$ 
        \STATE Calculate OT-based distribution matching loss in the P-space as $\mathcal{L}_{bv}$ in Eq.~\ref{eq5} and the orthogonal loss in Eq.~\ref{eq6}
        \STATE Generate Projection-Space mask $\Mmat$ by Eq.~\ref{eq4} and mask the P-space representations by $\Hmat^k = \Mmat^{k} \odot \Zmat$
        \STATE Reconstruct the multiple masked representations in parallel as $\Xmat^{rec}_k=\Phimat_D(\Hmat^k;\thetav_D)$
        \STATE Build distributions for association vectors $\piv$ and association prototypes $\Upsilon$ as $P(\piv) = \sum_{n=1}^{N} \frac{1}{N} \delta_{\piv_n}$ and $Q(\Upsilon) = \sum_{m=1}^{M} \frac{1}{M} \delta_{\gammav^m}$
        \STATE Calculate OT-based distribution matching loss of association prototypes as $\mathcal{L}_{ap}$ in Eq.~\ref{eq7} and the multiple reconstruction loss by $\mathcal{L}_{rec} = \frac{1}{K} \sum_{k=1}^K ||\Xmat - \Xmat_k^{rec}||^2_2$
        \STATE Update model parameters by minimizing $\mathcal{L} = \mathcal{L}_{rec} + \lambda_{bv}\cdot\mathcal{L}_{bv} + \lambda_{ap}\cdot\mathcal{L}_{ap} +\lambda_{orth}\cdot\mathcal{L}_{orth}$
    \ENDFOR    
\end{algorithmic}
\end{algorithm}

\section{Experiments}

\textbf{Datasets.} Following previous work~\cite{yin2024mcm}, we use 20 commonly used tabular datasets spanning multiple domains, including environmental studies, satellite remote sensing, healthcare, finance, etc. Specifically, 12 datasets were sourced from OOD \cite{Rayana:2016} and 8 from ADBench \cite{han2022adbench}. Detailed descriptions of these datasets are provided in the appendix \ref{sec:datasets characteristics}.

\textbf{Evaluation Metrics.} Following the methodology outlined in the literature \cite{zong2018deep, bergman2020classification}, we randomly selected one-half of the normal samples as the training set. The other half of the normal samples are combined with all the anomalous samples to form the test set. We adopted the Area Under the Receiver Operating Characteristic Curve (AUC-ROC) and the Area Under the Precision-Recall Curve (AUC-PR) as our evaluation metrics.

\textbf{Implementation Details.} Both the encoder and decoder contain an NPT architecture, each consisting of an ABD layer and an ABA layer with each attention module containing 4 attention heads. Following \cite{kossen2021self}, we utilize a Row-wise feed-forward (rFF) network containing one hidden layer, employing a 4x expansion factor and GeLU activation with the dropout rate of 0.1 for both attention weights and hidden layers. For input and output embeddings, the hidden size of the linear layer to encode the feature is set to 16. For each dataset, we use 5 basis vectors and 5 association prototypes. 
LAMB \cite{you2019large} with $\beta$ is used as the optimizer including a Lookahead \cite{zhang2019lookahead} wrapper with update rate $\alpha = 0.5$ and $k = 6$ steps between updates. In the first 10 epochs of training, we apply a warm-up strategy \cite{he2016deep} to gradually decrease the learning rate, followed by a cosine annealing strategy \cite{loshchilov2016sgdr} to adjust the learning rate in subsequent epochs. {The whole model is trained end-to-end under the loss $\mathcal{L}$, and $\lambda_{bv},\lambda_{ap}$ and $\lambda_{orth}$ is set to {1, 1, and 0.1}}. Unless specified otherwise, we set the hyper-parameter of regularized entropy as $\lambda = 0.1$, and score weights $\kappa$ and $\alpha$ are set to 0.01. 


\textbf{Baselines.} We extensively compare our model with tabular anomaly detection methods including both traditional machine learning and deep learning approaches. The traditional machine learning methods include KNN~\cite{ramaswamy2000efficient}, IForest \cite{liu2008isolation}, LOF \cite{breunig2000lof}, OCSVM \cite{scholkopf1999support} and {GMM \cite{agarwal2007detecting}}, and the deep learning methods include DeepSVDD \cite{ruff2018deep}, GOAD \cite{bergman2020classification}, NeuTralAD \cite{qiu2021neural}, ICL \cite{shenkar2022anomaly}, {DTE-C \cite{livernoche2023diffusion}}, NPT-AD \cite{thimonier2024beyond}, and MCM \cite{yin2024mcm}. 
It is noteworthy that the comparison experiments are based on the comprehensive open-source libraries PYOD~\cite{zhao2019pyod} (reproduction of KNN IForest, LOF, OCSVM, GMM, and DeepSVDD) and DeepOD~\cite{xu2023deep,xu2024calibrated} (reproduction of GOAD, NeuTral, and ICL). The remaining baselines were implemented based on the official open-source code. Furthermore, we also compare the MCM model in combination with the NPT model for comparison. In our experiments, all methods were implemented using consistent dataset splits and preprocessing procedures in line with recent research \cite{qiu2021neural, shenkar2022anomaly, yin2024mcm}. We report the average performance over three runs throughout this paper.


\subsection{Main Result}

\begin{figure*}[ht]
    \centering
    \begin{minipage}{0.49\textwidth}
        \centering
        \includegraphics[width=\linewidth]{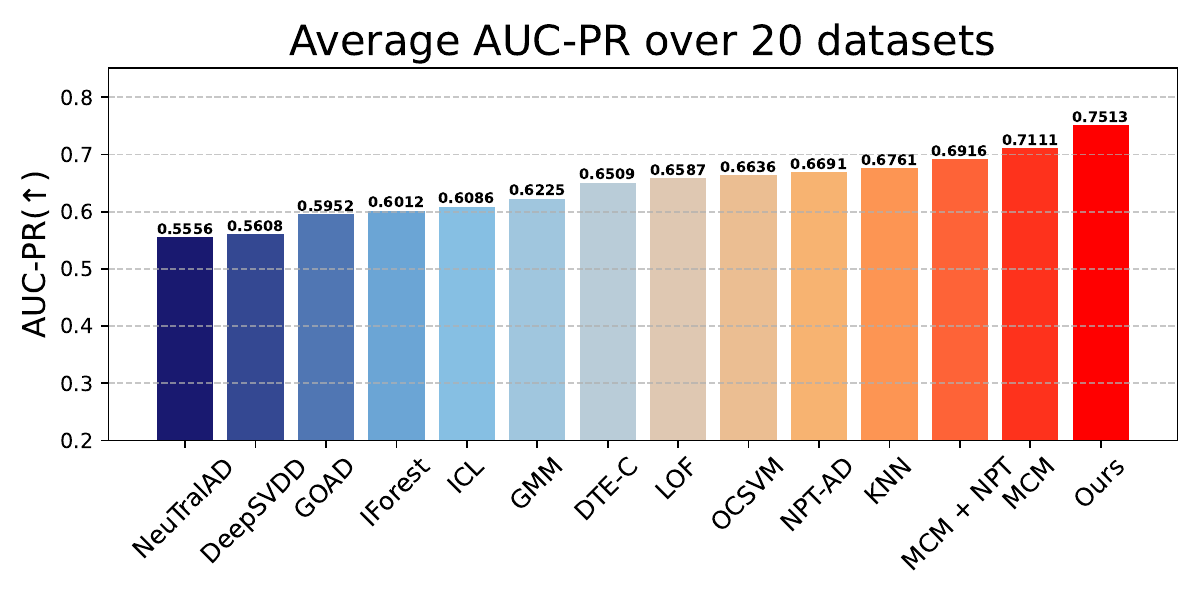}
        \caption{{AUC-PR of models over 20 datasets ($\uparrow$).}}
        \label{fig: Average AUC-PR figure}
    \end{minipage}
    \hfill
    \begin{minipage}{0.49\textwidth}
        \centering
        \includegraphics[width=\linewidth]{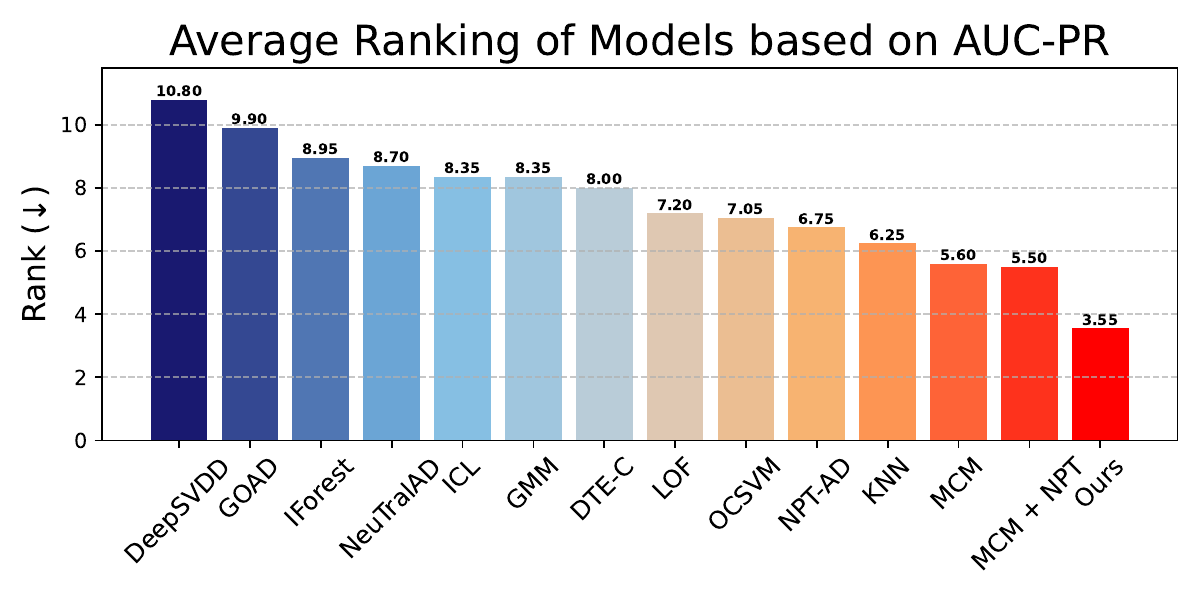}
        \caption{{Ranking of model based on AUC-PR ($\downarrow$).}}
        \label{fig: Average PR Ranking figure}
    \end{minipage}
\end{figure*}

\begin{figure*}[ht]
    \centering
    \begin{minipage}{0.48\textwidth}
        \centering
        \includegraphics[width=\linewidth]{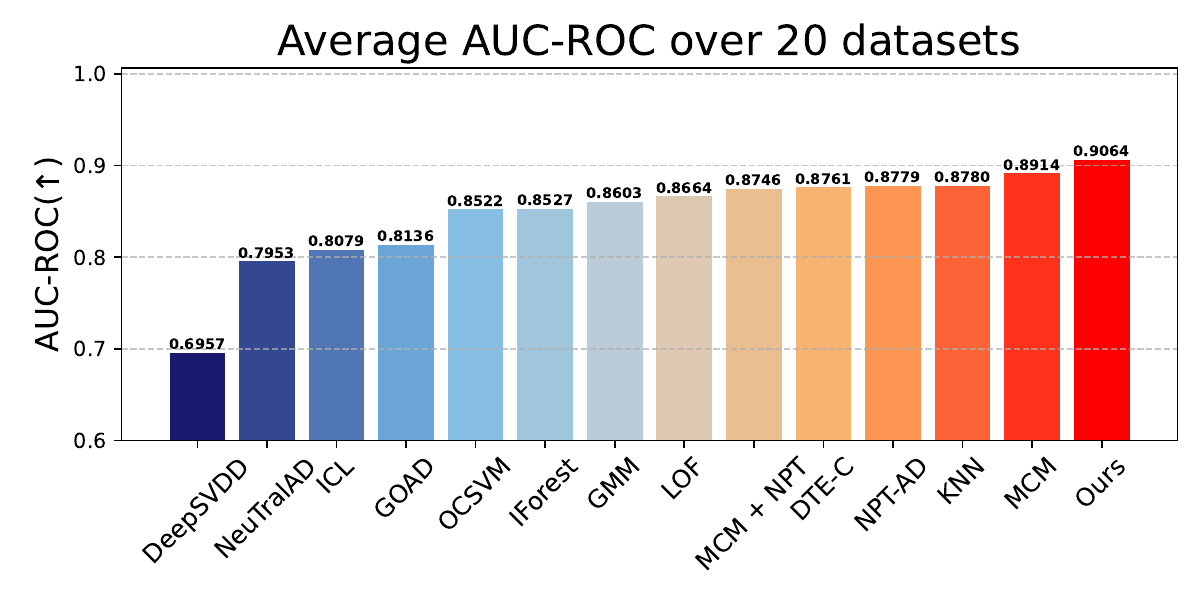}
        \caption{{AUC-ROC of models over 20 datasets ($\uparrow$).}}
        \label{fig: Average AUC-AUC figure}
    \end{minipage}
    \hfill
    \begin{minipage}{0.48\textwidth}
        \centering
        \includegraphics[width=\linewidth]{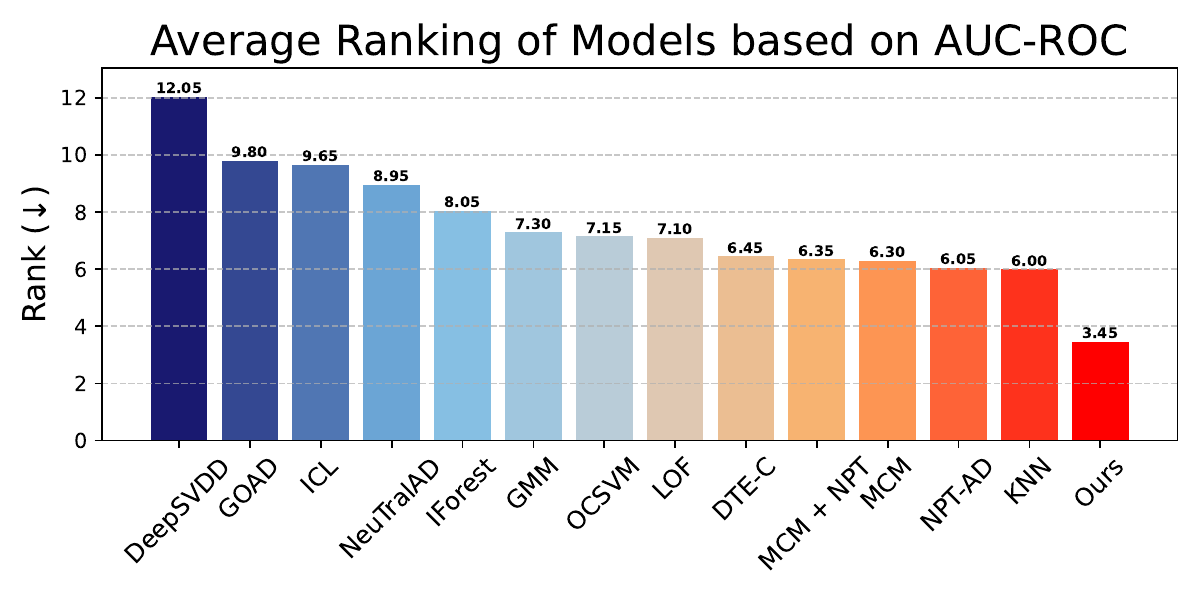}
        \caption{{Ranking of model based on AUC-ROC($\downarrow$).}}
        \label{fig: Average AUC Ranking figure}
    \end{minipage}
\end{figure*}

The AUC-PR and AUC-ROC results of our method and the other competitors are respectively shown in Fig.~\ref{fig: Average AUC-PR figure} and Fig.~\ref{fig: Average AUC-AUC figure}. The average ranking results are also shown in Fig.~\ref{fig: Average PR Ranking figure} and Fig.~\ref{fig: Average AUC Ranking figure}, while detailed results on each datasets are listed in Table~\ref{tab:Comaprison of AUC-PR} and Table~\ref{tab:Comaprison of AUC-ROC}, which shows that our method achieves the competitive performances over all datasets. Notably, our method significantly outperforms other methods on several datasets, {such as Optdigits and Wbc, leading to 8.12\% and 6.56\% improvements respectively}. Even in cases where our method slightly falls short of the best-performing method, its performance remains commendable, with performance gaps within acceptable ranges. On average, our method achieves around 4\% improvement over the second-best comparison method MCM, which demonstrates the effectiveness of our proposed method. The attempt to incorporate the deviation with normal patterns for tabular anomaly detection is feasible. Due to space limitations, we only present AUC-PR results in this table. Furthermore, the comparison results of AUC-ROC are listed in Table~\ref{tab:Comaprison of AUC-ROC}, in which the overall trends are consistent with those of AUC-PR. As for average results, our approach outperforms all the others and achieves the best or second-best performance on {14 out of the 20 datasets on AUC-ROC.} These evaluation results demonstrate the effectiveness and the generalizability of our model for tabular anomaly detection. 

\begin{table*}
    \centering
    \caption{{Comparison results of AUC-PR on 20 datasets. The best are bold and the second best are underlined.}}
    \label{tab:Comaprison of AUC-PR}
    \tabcolsep=0.08cm
    \resizebox{\textwidth}{!}{
    \begin{tabular}{l | c c c c c c c c c c c c c | c}
    \toprule
        Dataset          & KNN     & IForest & LOF     & OCSVM   & {GMM}      & DeepSVDD         & GOAD            & NeuTralAD       & ICL              & {DTE-C}            & NPT-AD           & MCM~            & MCM + NPT        & Ours          \\ \midrule
Arrhythmia                               & 0.6185                      & \textbf{0.6759}                 & 0.6031          & 0.6336          & 0.5247 & 0.5515          & 0.5988 & 0.5817          & 0.5773          & \underline{0.6609}    & 0.4779          & 0.5765       & 0.5956          & 0.6164          \\
Breastw                                  & 0.9908                      & 0.9945                          & 0.9315          & 0.9846          & 0.9826 & 0.9024          & 0.9782 & 0.6866          & 0.9459          & 0.9207          & 0.9815          & 0.9902       & \textbf{0.9976} & \underline{0.9973}    \\
Campaign                                 & 0.4898                      & 0.3995                          & 0.4962          & 0.4962          & 0.5157 & 0.3302          & 0.4518 & 0.3867          & 0.4291          & 0.4877          & 0.4852          & \underline{0.5543} & 0.4954          & \textbf{0.5826} \\
Cardio                                   & 0.7344                      & 0.8045                          & 0.6471          & \underline{0.8484}    & 0.7678 & 0.7269          & 0.8405 & \textbf{0.8570} & 0.8054          & 0.7125          & 0.8216          & 0.8432       & 0.8352          & 0.8445          \\
Cardiotocography & 0.6655                      & 0.6135                          & 0.7089          & \textbf{0.7291} & 0.6525 & 0.5976          & 0.6840 & 0.6238          & 0.6443          & 0.5462          & 0.6443          & 0.7007       & \underline{0.7108}    & 0.6962          \\
Census                                   & 0.2132                      & 0.1266                          & 0.1428          & 0.2074          & 0.2011 & 0.0756          & 0.1148 & 0.1206          & 0.1850          & 0.1758          & \underline{0.2363}    & 0.2337       & 0.2121          & \textbf{0.2970} \\
Fraud                                    & 0.2511                      & 0.0153                          & 0.3517          & 0.3398          & 0.6122 & \textbf{0.7627} & 0.5076 & 0.4730          & 0.5909          & 0.6343          & 0.3972          & 0.5884       & \underline{0.6389}    & 0.5377          \\
Glass                                    & 0.1614                      & 0.1187                          & 0.2452          & 0.1378          & 0.1362 & 0.1637          & 0.1205 & 0.1873          & 0.2296          & \textbf{0.5002} & 0.2235          & 0.1752       & 0.2414          & \underline{0.3880}    \\
Ionosphere                               & 0.9759                      & 0.9164                          & 0.9469          & 0.9737          & 0.9722 & 0.8636          & 0.9484 & \underline{0.9818}    & 0.9771          & 0.9801          & \textbf{0.9875} & 0.9740       & 0.9803          & 0.9813          \\
Mammography                              & 0.4167                      & 0.4263                          & 0.2973          & 0.4211          & 0.4276 & 0.0429          & 0.1614 & 0.0387          & 0.1792          & 0.4045          & 0.4133          & 0.3173       & \textbf{0.4778} & \underline{0.4398}    \\
NSL-KDD          & 0.7508                      & 0.8189                          & 0.7515          & 0.7553          & 0.8119 & 0.4876          & 0.8536 & 0.8676          & 0.5621          & \textbf{0.8932} & 0.8603          & 0.8572       & 0.8237          & \underline{0.8823}    \\
Optdigits                                & 0.3446                      & 0.1426                          & 0.5301          & 0.6977          & 0.1484 & 0.1103          & 0.0847 & 0.1736          & 0.4400          & 0.1465          & 0.1251          & \underline{0.7135} & 0.3577          & \textbf{0.7957} \\
Pima                                     & \textbf{0.9772}             & 0.5185                          & \underline{0.8318}    & 0.6972          & 0.1380 & 0.6409          & 0.6618 & 0.6081          & 0.6462          & 0.6075          & 0.6858          & 0.6759       & 0.7205          & 0.7308          \\
Pendigits                                & 0.7078                      & 0.7185                          & 0.6578          & 0.4914          & 0.6972 & 0.2161          & 0.3319 & 0.5777          & 0.4003          & 0.4712          & \textbf{0.9388} & 0.7338       & 0.7663          & \underline{0.9260}    \\
Satellite                                & \underline{0.8925}                & 0.8320                          & 0.8793          & 0.8227          & 0.8485 & 0.8401          & 0.8077 & 0.8654          & \textbf{0.8976} & 0.8496          & 0.8540          & 0.8502       & 0.8356          & 0.8433          \\
Satimage-2                               & 0.9782                      & 0.9217                          & 0.9311          & 0.9732          & 0.7537 & 0.7106          & 0.8625 & 0.8367          & 0.8599          & 0.7473          & \textbf{0.9859} & 0.9792       & 0.9642          & \underline{0.9844}    \\
Shuttle                                  & 0.9751                      & 0.9815                          & \textbf{0.9939} & 0.9732          & 0.9687 & \underline{0.9875}    & 0.9765 & 0.9804          & 0.9766          & 0.9430          & 0.9656          & 0.9666       & 0.9165          & 0.9377          \\
Thyroid                                  & \underline{0.8243}                & 0.7427                          & 0.6693          & 0.761           & 0.7557 & 0.5502          & 0.7292 & 0.8095          & 0.6834          & \textbf{0.8338} & 0.7851          & 0.8188       & 0.5385          & 0.7685          \\
Wbc                                      & 0.7832                      & 0.7649                          & \underline{0.8080}    & 0.7403          & 0.7503 & 0.7535          & 0.7292 & 0.2130          & 0.7795          & 0.6453          & 0.7497          & 0.7466       & 0.7973          & \textbf{0.8451} \\
Wine                                     & 0.7704                      & 0.4916                          & 0.7511          & 0.5878          & 0.7850 & 0.9021          & 0.4608 & 0.2425          & 0.3631          & 0.8573          & 0.7635          & \underline{0.9269} & 0.9260          & \textbf{0.9323} \\ \midrule
Average                                  & 0.6761                      & 0.6012                          & 0.6587          & 0.6636          & 0.6225 & 0.5608          & 0.5952 & 0.5556          & 0.6086          & 0.6509          & 0.6691          & \underline{0.7111} & 0.6916          & \textbf{0.7513} \\ \bottomrule

    \end{tabular}
    }
\end{table*}

\begin{table*}
    \centering
    \caption{\footnotesize{{Comparison results of AUC-ROC on 20 datasets. {The best are bold and the second best are underlined.}}}}
    \label{tab:Comaprison of AUC-ROC}
    \tabcolsep=0.08cm
    \resizebox{\textwidth}{!}{
    \begin{tabular}{l | c c c c c c c c c c c c c | c} 
    \toprule
        Dataset          & KNN             & IForest         & LOF     & OCSVM           & {GMM}       & DeepSVDD & GOAD            & NeuTralAD       & ICL              & {DTE-C}            & NPT-AD                   & MCM~             & MCM + NPT        & Ours          \\ \midrule
Arrhythmia                               & 0.8070                      & \underline{0.8362}                    & 0.8030          & 0.7998 & 0.7881          & 0.7502       & 0.8146       & 0.8192       & 0.7937          & \textbf{0.8816} & 0.7103          & 0.7826          & 0.8012          & 0.8147          \\
Breastw                                  & 0.9910                      & 0.9946                          & 0.9655          & 0.9863 & 0.9849          & 0.7700       & 0.9766       & 0.7987       & 0.9622          & 0.9360          & 0.9834          & 0.9911          & \textbf{0.9977} & \underline{0.9973}    \\
Campaign                                 & 0.7695                      & 0.6976                          & 0.7762          & 0.7746 & 0.7983          & 0.3951       & 0.7201       & 0.6353       & 0.7032          & 0.7995          & 0.7778          & \underline{0.8619}    & 0.7830          & \textbf{0.8693} \\
Cardio                                   & 0.9112                      & 0.9508                          & 0.9085          & 0.9618 & 0.9310          & 0.7508       & \underline{0.9639} & 0.9601       & 0.9140          & 0.8889          & 0.9211          & 0.9635          & 0.9543          & \textbf{0.9653} \\
Cardiotocography & 0.7334                      & 0.7475                          & 0.8116          & 0.8157 & 0.7493          & 0.6432       & 0.7670       & 0.6799       & 0.6840          & 0.6181          & 0.6840          & 0.8024          & \underline{0.8207}    & \textbf{0.8210} \\
Census                                   & 0.7201                      & 0.5862                          & 0.6022          & 0.7029 & 0.7061          & 0.5431       & 0.5330       & 0.4986       & 0.6725          & 0.6834          & 0.7008          & \underline{0.7515}    & 0.7212          & \textbf{0.7622} \\
Fraud                                    & 0.9536                      & 0.8312                          & 0.9562          & 0.9563 & \textbf{0.9577} & 0.8846       & 0.9356       & 0.8892       & 0.9143          & 0.9413          & \underline{0.9564}    & 0.9025          & 0.8840          & 0.9531          \\
Glass                                    & 0.7713                      & 0.6882                          & 0.7573          & 0.6915 & 0.6721          & 0.6375       & 0.6257       & 0.7907       & 0.8196          & \textbf{0.9390} & 0.7843          & 0.7480          & 0.7118          & \underline{0.8353}    \\
Ionosphere                               & 0.9704                      & 0.9144                          & 0.9383          & 0.9645 & 0.9609          & 0.9105       & 0.9366       & \underline{0.9776} & 0.9741          & 0.9713          & \textbf{0.9805} & 0.9621          & 0.9724          & 0.9738          \\
Mammography                              & 0.8746                      & 0.8827                          & 0.8255          & 0.8904 & 0.8802          & 0.4807       & 0.4527       & 0.4604       & 0.5653          & 0.8680          & \underline{0.8928}    & 0.8660          & \textbf{0.9078} & 0.8882          \\
NSL-KDD          & 0.5760                      & 0.7888                          & 0.5806          & 0.5824 & 0.6780          & 0.4953       & \underline{0.8524} & 0.7521       & 0.2665          & 0.8509          & 0.8126          & \textbf{0.9606} & 0.6785          & 0.8513          \\
Optdigits                                & 0.9516                      & 0.8169                          & 0.9774          & 0.6326 & 0.8360          & 0.6052       & 0.6936       & 0.7743       & 0.9552          & 0.8254          & 0.8084          & \textbf{0.9837} & 0.9537          & \underline{0.9825}    \\
Pima                                     & \textbf{0.9986}             & 0.9607                          & \underline{0.9911}    & 0.6983 & 0.8377          & 0.5861       & 0.6816       & 0.6238       & 0.6231          & 0.6124          & 0.7161          & 0.6503          & 0.7368          & 0.7234          \\
Pendigits                                & 0.7373                      & 0.7291                          & 0.7082          & 0.9628 & 0.7242          & 0.3064       & 0.9297       & 0.9720       & 0.8334          & 0.9769          & \textbf{0.9983} & 0.9906          & 0.9750          & \underline{0.9961}    \\
Satellite                                & \underline{0.8760}                & 0.7773                          & 0.8405          & 0.7575 & 0.8030          & 0.5848       & 0.7356       & 0.8204       & \textbf{0.8805} & 0.7932          & 0.7914          & 0.7949          & 0.7827          & 0.7992          \\
Satimage-2                               & 0.9989                      & 0.9933                          & 0.9967          & 0.997  & 0.9942          & 0.7068       & 0.9881       & 0.9954       & 0.9828          & 0.9951          & \textbf{0.9995} & 0.9987          & 0.9989          & \textbf{0.9995} \\
Shuttle                                  & 0.9990                      & 0.9956                          & \textbf{0.9997} & 0.9966 & 0.9955          & \underline{0.9995} & 0.9931       & 0.9950       & 0.9889          & 0.9976          & 0.9986          & 0.9986          & 0.9940          & 0.9965          \\
Thyroid                                  & 0.9870                      & \underline{0.9893}                    & 0.9577          & 0.983  & 0.9750          & 0.9476       & 0.9680       & 0.9881       & 0.9223          & \textbf{0.9896} & 0.9762          & 0.9636          & 0.9310          & 0.9750          \\
Wbc                                      & 0.9606                      & 0.9633                          & 0.9617          & 0.9604 & 0.9577          & 0.9340       & 0.9154       & 0.7364       & 0.9087          & \underline{0.9681}    & 0.9577          & 0.9510          & 0.9617          & \textbf{0.9737} \\
Wine                                     & 0.9733                      & 0.9100                          & 0.9700          & 0.93   & 0.9767          & \underline{0.9833} & 0.7883       & 0.7383       & 0.7927          & \textbf{0.9864} & 0.9567          & 0.9037          & 0.9260          & 0.9507          \\ \midrule
Average                                  & 0.8780                      & 0.8527                          & 0.8664          & 0.8522 & 0.8603          & 0.6957       & 0.8136       & 0.7953       & 0.8079          & 0.8761          & 0.8779          & \underline{0.8914}    & 0.8746          & \textbf{0.9064} \\ \bottomrule

    \end{tabular}}
\end{table*}

\subsection{Ablation Study}
In this section, we explore the effectiveness of different components within PTAD. The average AUC-ROC and AUC-PR across all datasets are reported in Table~\ref{tab:ablation}. The variations and observations are listed as follows:

\textbf{i) Data-space Masking}: Incorporating the data-space mask strategy into the baseline model solely composed of two NPT layers leads to 1.27\% improved AUC-PR, demonstrating the effectiveness of soft mask modeling in data space.

\textbf{ii) Single P-space Learnable Masking}: We generate a single P-space mask and decode masked representation via a single-branch decoder, which results in 3.4\% performance gain on AUC-PR, indicating incorporating learnable masks in P space contributes to modeling the global correlation within tabular data. 

\textbf{iii) Multiple P-space Learnable Masking}: Multiple masks and multiple-branch decoding lead to 4.89\% AUC-PR improvement, highlighting the necessity of our multiple designations.

\textbf{iv) Random Masking}: We randomly generate masks with the same masking rate as our multiple learnable masks. Without the guidance of the basis vectors for generating the mask, the performance degrades by 5.02\%, which further confirms the important role of our learnable mask generation method.

\textbf{v) Association Prototypes}: Incorporating the association prototypes leads to 5.72\%, which facilitates us to evaluate the extent of the abnormality by measuring the deviations. 

\textbf{vi) Orthogonality Constrain}: We further validate the orthogonality constrain of basis vectors, the performance gap showcases the orthogonality contributes to anomaly detection by forming disentangled features.

Overall, the comprehensive version performs best, demonstrating its effectiveness and efficiency as a harmonious combination of its components. 

\begin{table*}[h]
\centering
\caption{{Ablation study of our method.}}
\label{tab:ablation}
\tabcolsep=0.08cm
\resizebox{1.0\textwidth}{!}{
\begin{tabular}{c| c| c| c| c| c| c| c}
\hline\hline
\multicolumn{4}{c|}{Masks} & & & & \\ \cline{1-4}
Data-space Mask &Single Learnable Mask & Multiple Learnable Masks  & Random Mask & Association Prototypes & Orthogonality Constrain  & AUC-PR & AUC-ROC \\ \hline
\rowcolor{red!20} $\times$           & $\times$       & $\times$          &  $\times$               &  $\times$     & $\times$  & 0.6381  & 0.8457   \\ 
\rowcolor{blue!20} \checkmark      & $\times$      & $\times$      &  $\times$                &  $\times$     & $\times$            & 0.6508& 0.8378  \\
\rowcolor{blue!20} \checkmark    &  \checkmark   & $\times$  & $\times$     &  $\times$     &  $\times$      & 0.6848& 0.8835 \\ 
\rowcolor{blue!20} \checkmark    & $\times$   & \checkmark   & $\times$      &  $\times$              
& \checkmark     & 0.7337& 0.8893 \\ 
\rowcolor{blue!20} \checkmark      &  $\times$         & $\times$    & \checkmark      &  $\times$     & $\times$            & 0.6835& 0.8766  \\
\rowcolor{green!20} \checkmark     & $\times$  &  $\times$     & $\times$     & \checkmark     &$\times$    & 0.7080 &0.8884\\ 
\rowcolor{yellow!20} \checkmark    & $\times$  & \checkmark &$\times$      & \checkmark  & $\times$   & 0.7392  & 0.8982\\ 
\rowcolor{gray!20} \checkmark    & $\times$ & \checkmark  &$\times$     & \checkmark  & \checkmark     & \textbf{0.7513}& \textbf{0.9064}\\ 
\hline\hline
\end{tabular}}
\end{table*}

\subsection{Robustness to Different Types of Anomalies}
Although the true distribution of anomalous samples is challenging to capture, previous works~\cite{steinbuss2021benchmarking,han2022adbench} have identified four common types of anomalies and proposed methods for generating normal and abnormal samples from real datasets. We follow \cite{han2022adbench} to generate data from the cardiotocography dataset and examine the robustness of our method encountering different anomalous: 

\begin{itemize}
\item \textbf{Local anomalies:} The classic GMM procedure~\cite{milligan1985algorithm, steinbuss2021benchmarking} is used to generate normal samples, after which a covariance scaling parameter $\alpha = 5$ is used to generate anomalous samples.
\item \textbf{Global anomalies:} The global anomalies are generated from a uniform distribution $ Unif(\alpha \cdot \min(\mathbf{X}^k), \alpha \cdot \max(\mathbf{X}^k))$, where the boundaries are defined as the $\min$ and $\max$ of an input feature, such as k-th feature $\mathbf{X}^k$. The hyperparameter $\alpha$ is established at 1.1, influencing the level of deviation exhibited by the anomalies.
\item \textbf{Dependency anomalies:} For generating independent types of anomalies, Vine Copula method~\cite{aas2009pair} is utilized to model the dependency structure of the original data, whereby the probability density function of the generated anomalies is established as completely independent by eliminating the modeled dependencies, which could refer to~\cite{martinez2014fault}. We use Kernel Density Estimation(KDE)~\cite{hastie2009elements} to estimate the probability density function of features and generate normal samples.
\item \textbf{Clustered anomalies:} We scale the mean feature vector of normal samples by $\alpha = 5$, such as $\hat{\mu} = \alpha\hat{\mu}$. The hyperparameter $\alpha$ scales GMM, controlling the distance between anomaly clusters and the normal for generating anomalies.
\end{itemize}

\begin{table*}[ht]
\centering
\caption{{Results of four types of anomalies generated from the cardiotocography dataset.}}
\label{tab:four anomalies results}
\resizebox{\textwidth}{!}{%
\begin{tabular}{c c c c c c c c c c c c c c}
\toprule
\textbf{Category} & \textbf{Metrics} & \textbf{KNN} & \textbf{IForest} & \textbf{LOF} & \textbf{OCSVM}  & \textbf{DeepSVDD} & \textbf{GOAD} & \textbf{NeuTralAD} & \textbf{ICL} & \textbf{NPT-AD} & \textbf{MCM} & \textbf{Ours} \\ \midrule
\multirow{2}{*}{\textbf{Local}}      & AUC-PR  & 0.2479 & 0.2489 & 0.2611 & 0.2593  & 0.3362 & 0.4797 & 0.4554 & 0.3456 & 0.3326 & 0.5002 & \textbf{0.5277} \\ 
& AUC-ROC & 0.8804 & 0.8657 & 0.8888 & 0.8779  & 0.7849 & 0.8809 & 0.8842 & 0.7138 & 0.4855 & \textbf{0.9009} & 0.8907 \\ \midrule
\multirow{2}{*}{\textbf{Global}}     & AUC-PR  & 0.3075 & 0.3272 & 0.3009 & 0.3168  & 0.4078 & 0.4663 & 0.4749 & 0.2635 & 0.4731 & 0.5775 & \textbf{0.6143} \\ 
& AUC-ROC & 0.9087 & 0.9157 & 0.9083 & 0.9120  & 0.8823 & 0.9227 & 0.9211 & 0.7104 & 0.9158 & 0.9441 & \textbf{0.9464} \\ \midrule
\multirow{2}{*}{\textbf{Dependency}} & AUC-PR  & 0.1895 & 0.1131 & 0.2318 & 0.1160  & 0.2361 & 0.1664 & 0.3350 & 0.2799 & 0.3467 & 0.4495 & \textbf{0.4879} \\ 
& AUC-ROC & 0.8567 & 0.7417 & 0.8918 & 0.7432  & 0.6702 & 0.6773 & 0.8628 & 0.8033 & 0.8364 & 0.9171 & \textbf{0.9239} \\ \midrule
\multirow{2}{*}{\textbf{Cluster}}    & AUC-PR  & 0.135  & 0.3298 & 0.0718 & 0.2797  & 0.1957 & 0.4748 & 0.1539 & 0.1718 & 0.2021 & 0.5316 & \textbf{0.5806} \\ 
& AUC-ROC & 0.7136 & 0.9136 & 0.4837 & 0.9088  & 0.6806 & 0.9157 & 0.5923 & 0.5266 & 0.6403 & 0.9232 & \textbf{0.9391} \\
\bottomrule
\end{tabular}
}
\end{table*}

We list the experimental results across the four types of anomalies in Table~\ref{tab:four anomalies results}. It can be seen that our model performs well across all four types of anomalies. Especially for dependency and cluster anomalies, our model significantly outperforms the second-best approach, showcasing our model's ability to distinguish these anomalies from normal samples. This might contribute to our special modeling for the dependencies across datapoints and features by multiple masking strategies and the record of both normal features and association patterns.

\begin{table}
    \centering
    \caption{{Computational cost on Campaign dataset}}
    \label{tab:computational cost comparison}
    \resizebox{0.45\textwidth}{!}{
    \begin{tabular}{l c c c c }
        \toprule
        & MCM &   NPT-AD   &  Our method    \\
        \midrule
        Params(M)                  &  \textbf{0.23}   &    \underline{2.97}   &    \underline{2.97}         \\
        Training Time(ms)          &  \textbf{23.35}  &  3384.77  &    \underline{130.20}       \\
        Inference Time(ms)         &  \textbf{5.87}   & 30904.05  &    \underline{22.58}        \\
        AUC-PR                     &  \underline{0.5543} &  0.4770   &\textbf{0.5826}  \\
        AUC-ROC                    &  \underline{0.8619} &  0.7915   &\textbf{0.8693}  \\
        \bottomrule
    \end{tabular}} 
\end{table}

\subsection{Discussion}
\textbf{Computational Cost.} Assessing computational cost is crucial for understanding model efficiency and feasibility in practical applications. We present a comprehensive comparison of the computational costs for our method with two strong competitors MCM \cite{yin2024mcm} and NPT-AD \cite{thimonier2024beyond}. 
Four metrics are reported with the average performance across ten times running.
As shown in Table~\ref{tab:computational cost comparison}, we report the computational cost and performances of different models on the campaign dataset. Our method could achieve strong performance while maintains a decent balance with computational cost. 

\begin{figure*}[ht]
    \centering
    \begin{minipage}{0.48\textwidth}
        \centering
        \includegraphics[width=\linewidth]{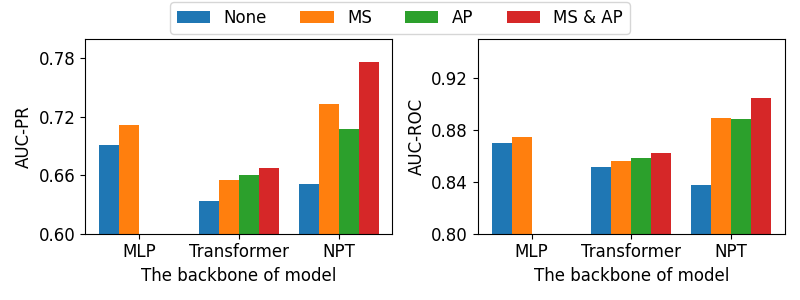}
        \caption{{Performance of various components conducted with different backbones: MLP, Transformer, NPT.}}
        \label{fig:Performance with different backbones}
    \end{minipage}
    \hfill
    \begin{minipage}{0.48\textwidth}
        \centering
        \includegraphics[width=\linewidth]{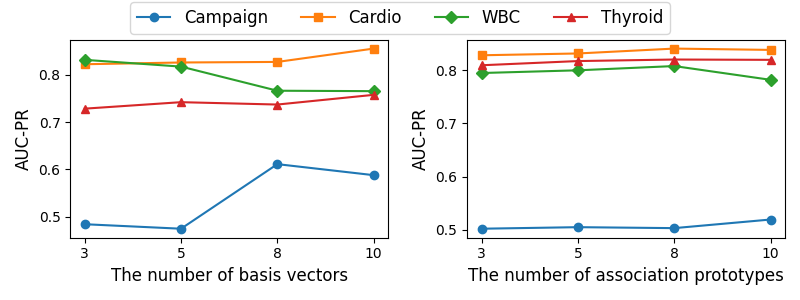}
        \caption{{Performance with different numbers of basis vectors (left) and association prototypes (right) on various datasets.}}
        \label{fig:Performance with different numbers}
    \end{minipage}
\end{figure*}
\textbf{Different Backbones.} The proposed masking strategy and association prototype learning is model-agnostic and flexible, which could be plug-and-played to the other backbone models, such as MLP and vanilla transformers. For MLP, we show the improvement achieved by the masking strategy. For models containing attention maps, such as vanilla transformer and NPT, we demonstrate the performance improvements brought by the two modules respectively and jointly. As illustrated in Fig.~\ref{fig:Performance with different backbones}, the masking strategy (MS) improves model performance across different backbones. For models with attention, i.e. Transformer and NPT, incorporating the association prototype (AP) of normal samples further enhances the model's ability to distinguish anomalous from normal samples. Furthermore, it is validated that jointly performing both strategies leads to the best performance. 

\textbf{Different Number of Basis Vectors \& Association Prototypes.} Fig.~\ref{fig:Performance with different numbers} illustrates the impact of varying numbers of basis vectors for masking and association prototypes on four datasets: Campaign, Cardio, WBC, and Thyroid, varying in sample numbers and feature dimensions. 
It can be seen that as the number of basis vectors increases, the performance across all datasets generally shows an increasing trend. As for the association prototypes, the performance is relatively stable with a slightly increasing trend as the number grows. Therefore, there remains a trade-off between the performance and the number of basis vectors and association prototypes, as a larger number indicates the increased computational cost. Therefore, to achieve a good balance between computational cost and performance, we set the number of the basis vectors and the association prototypes as 5 in our experiments, aiming to achieve better performances with relatively low costs.


\begin{table}
    \centering
    \caption{{Performances on AUC-PR/AUC-ROC with MSE or OT distances to learn basis vectors and association prototypes.}}
    \label{tab:Different loss performance on AUC-PR}
    \resizebox{0.49\textwidth}{!}{
    \begin{tabular}{c c | c | c }
        \hline \hline
        & &\multicolumn{2}{c}{Association Prototypes} \\ \cline{3-4}
        \multicolumn{2}{c|}{AUC-PR/AUC-ROC} & MSE     & OT    \\ \hline
        \multirow{2}{*}{Basis Vectors}
        &MSE         & 0.6306 / 0.8448 &  0.6543 / 0.8607   \\
        &OT          & 0.6649 / 0.8696   &  \textbf{0.7513 / 0.9064}   \\
        \hline \hline
    \end{tabular}}
\end{table}

\textbf{Distance Measurement.} For discussing the distance metric for learning basis vectors and association prototypes, we report the results utilizing MSE distance or OT distance in Table~\ref{tab:Different loss performance on AUC-PR}. It can be seen that OT distance facilitates optimizing both the basis vector and the association prototype, leading to a large improvement compared to MSE distance. This might contribute to our OT-based method views the points-to-points distance between discrete representations as a transport calibration between two distributions, making a smooth transport and appropriate measurement. 

\textbf{Different weights of orthogonal loss and anomaly scores.} \label{sec:Different weight for orthogonal loss and prototype score}
To investigate the influence of the orthogonal loss and anomaly scores, we illustrate the impact of different weights on AUC-PR and AUC-ROC across four datasets. 

\begin{figure}[ht]
    \centering
    \begin{minipage}{0.48\textwidth}
        \centering
        \includegraphics[width=\linewidth]{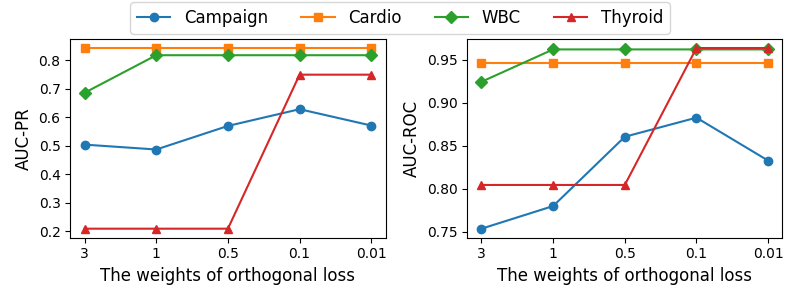}
        \caption{{Comparison Results of different weights of orthogonal constrain}}
        \label{fig:The weights of orthogonal loss}
    \end{minipage}
    \hfill
    \begin{minipage}{0.48\textwidth}
        \centering
        \includegraphics[width=\linewidth]{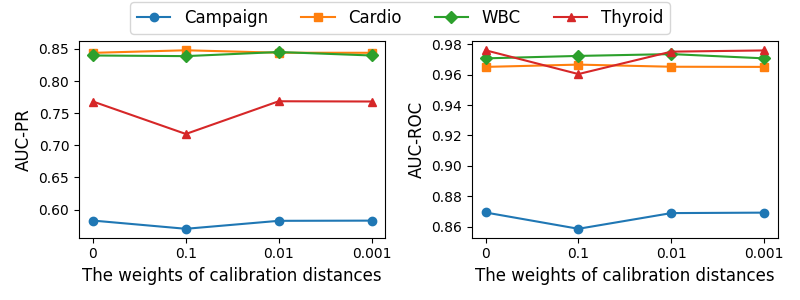}
        \caption{{Comparison Results of different weights of calibration distances}}
        \label{fig:The weights of prototype score}
    \end{minipage}
\end{figure}

1) For the orthogonal loss, we conduct experiments and illustrate the results in Fig.~\ref{fig:The weights of orthogonal loss}. It can be seen that the performance is stable on WBC and Cardio datasets while sensitive on the other datasets, which might be due to the tradeoff between regularizing the basis vectors to disentangle through orthogonal loss and ensuring an accurate representation of the P-space representation of normal data. To balance two aspects, we choose 0.1 as the weight of the orthogonal loss in our experiments.

2) Regarding the anomaly score, Fig.~\ref{fig:The weights of prototype score} displays the results over different weights across four datasets, where we set the same coefficients $\kappa$ and $\alpha$ of calibration distance. It can be seen that the performance is insensitive to the weights of both anomaly scores $\sv^{bv}$ and $\sv^{ap}$. To ensure stability in the anomaly detection, we choose 0.01 as our weighting coefficient. By evaluating the average performance of 20 datasets, the calibration distances complement the anomaly score and further enhance model performance.

\begin{figure*}[th]
\centering
\includegraphics[width=1.0\textwidth]{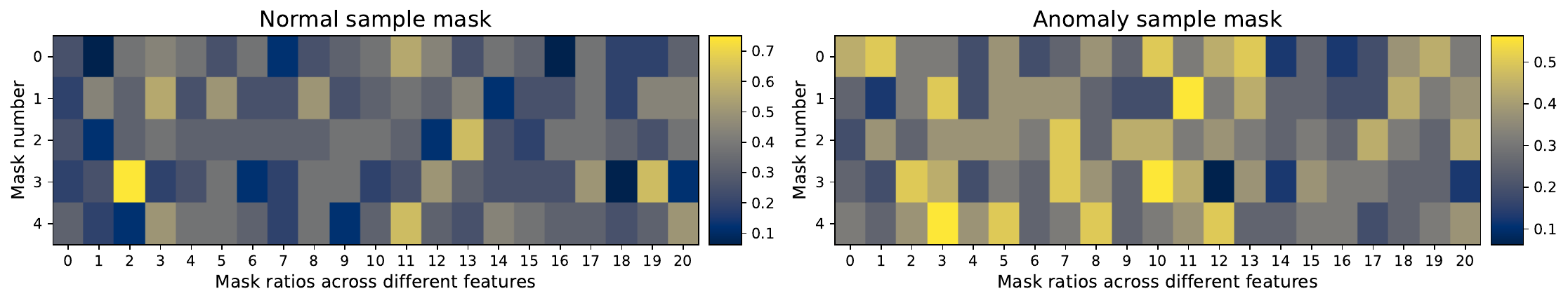}
\caption{{Visualization of P-space masks. The left figure corresponds to the normal sample, and the right figure refers to the abnormal sample, both possessing the same average masking rates.}}
\label{fig:Visualization of Projection-Space mask}
\end{figure*}

\textbf{Visualization of the P-space Masks.} To intuitively understand the P-space masking strategy, we calculate the masking rates of each corresponding feature for visual analysis. As shown in Fig.~\ref{fig:Visualization of Projection-Space mask}, we selected normal and anomalous sample masks of the Cardio test data with the same average masking rate, i.e. the rates of zeros in both masks are approximately 32\%. It can be observed that in the normal sample masks, a high masking rate is only observed for a subset of features, whereas the majority of features exhibit low masking rates, indicating that most of the normal samples are close to the basis vectors of P-space. In contrast, the masking patterns of abnormal samples are quite different, as most features have high masking rates, which illustrate deviations from the normal basis vectors and indicate anomalous. By masking those positions with higher possibilities of abnormality, the decoder is motivated to reconstruct the anomalous samples as normal outputs, leading to larger reconstruction errors for instructing anomalous.

\begin{figure*}[th!]
    \centering
        \includegraphics[width=\linewidth]{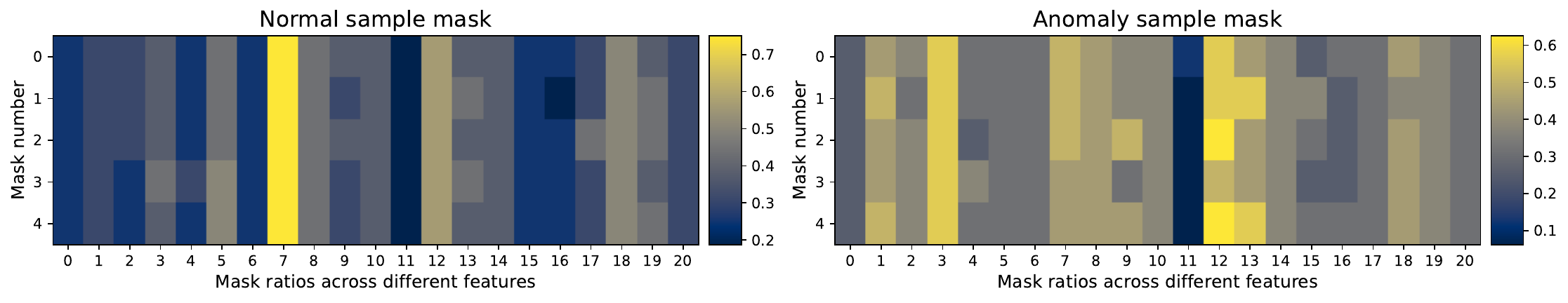}
        \caption{Visualization of the soft masks. The left figure corresponds to the normal sample, and the right figure refers to the abnormal sample, both possessing the same average masking rates.}
        \label{fig:soft mask visualization}
\end{figure*}

\textbf{Soft mask visualization.}
\label{sec:soft mask visualization}
To illustrate the soft masks, we visualize several randomly selected soft masks in Fig.~\ref{fig:soft mask visualization}. Note that the relationships across features are relatively regular in each dataset. The data-space masks try to find and automatically learn such regular patterns of relationships across features and embed them into the input. Specifically, some features are more critical for indicating anomalies while others are inconsequential. The soft masks could perform data-adaptive information bias to different features. As shown in Fig.~\ref{fig:soft mask visualization}, the soft masks are regular across different features. The data-space masks could uncover the statistical correlations between masked and unmasked positions across data points. 

Compared to the regular binary mask with a mask value of either 0 or 1, we apply the soft mask in raw data space with values between 0 and 1, providing a more flexible degree of information blocking and avoiding the complete lost of some features. When applying soft masks, the model can not only choose which features to mask, but also the degree of masking. This brings more flexibility to the model and is conducive to learning diverse and optimal masks under which the masked normal data can be reconstructed better than anomalies. Furthermore, compared to the random masking strategy which produces meaningless masks, our learnable masking strategy can not only choose which features to mask but also the degree of masking, generating optimal masks for our purpose. By reconstructing masked positions by capturing their correlations between unmasked positions, we train the soft masks to capture intrinsic correlations existing in the raw data space, and anomalies can be judged by whether deviating from such correlations.

\section{Conclusion}

In this paper, we propose an elaborately designed method PTAD, which is problem-oriented for tabular anomaly detection. We explore mask modeling and prototype learning to enhance anomaly detection performance. Masking modeling involves generating data-adaptive soft masks in the data space and multiple learnable masks in disentangled projection space with orthogonal basis vectors. The association prototypes are learned to extract normal characteristic correlations to capture the global cross-feature dependencies. Our model is derived from a distribution matching perspective and formulated as two optimal transport problems, where the calibration costs further refine the anomaly scoring function. Extensive quantitative and qualitative experimental results demonstrated our model's effectiveness, robustness, and generalizability. We hope that our way of modeling characteristic patterns of tabular data could give a new perspective on tabular anomaly detection and have the potential to be extended to wider fields of view.

\appendices
\section{{Training Pipeline}}
\label{sec:Training Pipeline}


{The training pipeline of PTAD consists of the following steps (Noting that we take the single vector as example, $\xv, \mv, \zv, \hv$ is referring to each row of the matrix $\Xmat, \Mmat, \Zmat, \Hmat$ ):}

{\textbf{Data-space Masking:} Following MCM~\cite{yin2024mcm}, for numerical features, we use their original scalar values; for categorical features, we use one-hot encoding to represent categorical features. Both the numerical features and one-hot categorical features are concatenated together as $\xv \in \Xmat \in \mathbb{R}^d$. Our data-space masks are data-adaptively learned for $d$ features as $\mv^{ds} \in \mathbb{R}^{d}$ by Eq.~\ref{eq3}. We then mask each feature by directly point-wise multiplicating the mask to its corresponding features as $\hat{\xv}=\xv \odot \mv^{ds}$. }

{\textbf{Encoding with an NPT Layer:}
The encoded representations of the masked data $\hat{\xv}$, are processed through learned linear layers to obtain the embedded representations of individual features. These feature embeddings are then passed into an NPT layer, which consists of an ABD layer followed by an ABA layer, resulting in the output $\zv \in \mathbb{R}^{H}$. Details can be found in Appendix C3 of NPT~\cite{kossen2021self}.}

{\textbf{Projection-Space Masking:} The P-space masks are generated by comparing the representation $\zv \in \mathbb{R}^{H}$ and each basis vector $\{\betav^k \in \mathbb{R}^{H}\}_{k=1}^K$ according to Eq.~\ref{eq4}, and generate $K$ masks $\{\mv^k \in \mathbb{R}^{H}\}_{k=1}^K$. Then we generate $K$ masked P-space representations $\{\hv^k = \mv^k \odot \zv \in \mathbb{R}^{H}\}_{k=1}^K$. Note we compute the OT loss by solving the OT problem between representation $\zv$ and $K$ basis vectors by Eq.~\ref{eq5}, which can be utilized to optimize the basis vectors.}

{\textbf{Decoding with an NPT Layer:} We parallelly input $K$ masked representations $ \{\hv^k\}_{k=1}^K $ into the decoder (an NPT layer consisting of an ABD and an ABA layer), respectively. The objective of the $K$ branch is the same: reconstructing the original tabular data. The architecture and parameters of the decoder are shared across $K$ branches. During decoding, we obtain the association vector $\piv \in \mathbb{R}^d$ and compute its OT distance with $M$ association prototypes $\{\gammav^m \in \mathbb{R}^d\}_{m=1}^M$ by Eq.~\ref{eq7}, which can be utilized to optimize the association prototypes.To obtain the estimated output features, we also use linear layers to transform the representations back into features, which serves as the inverse process of the input stage.}

{\textbf{Training Parameters:} In this training pipeline, the parameters need to be optimized including: NPT Encoder $\Phimat_E$, NPT Decoder $\Phimat_D$, basis vectors $\mathcal{B} = \{\betav^1,...,\betav^{K}\}_{k=1}^K $, association prototypes $\Upsilon = \{ \gammav^1,...,\gammav^M\}_{m=1}^M$, and $\{\Wmat_1,\Wmat_2,\Wmat_3\}$ of data-space soft mask generator.} 

\section{Detailed Datasets characteristics} \label{sec:datasets characteristics}
Table~\ref{tab:Datasets details} shows detailed information about the datasets utilized in our experiments, including the total number of samples, data dimensions, and the number of anomalous samples. These datasets include multiple domains, such as environmental studies, satellite remote sensing, healthcare, and so on, mainly sourced from OOD~\cite{Rayana:2016} and ADBench~\cite{han2022adbench}.

\begin{table}[h]
\belowrulesep=0pt
\aboverulesep=0pt
    \centering
    \caption{Details of 20 Datasets.}
    \label{tab:Datasets details}
    \resizebox{0.4\textwidth}{!}{
    \begin{tabular}{l|c c c}
        \toprule
        \text{Dataset}       & Samples & Dims & Anomaly \\
        \midrule
        Arrhythmia             & 452              & 274           & 66 (14.6\%)           \\
        Breastw                & 683              & 9             & 239 (35.0\%)          \\
        Campaign               & 41188            & 62            & 4640 (11.3\%)         \\
        Cardio                 & 1831             & 21            & 176 (9.6\%)           \\
        Cardiotocography       & 2114             & 21            & 466 (22.0\%)         \\
        Census                 & 299285           & 500           & 18568 (6.2\%)         \\
        Fraud                  & 284807           & 29            & 492 (0.2\%)           \\
        Glass                  & 214              & 9             & 9 (4.2\%)            \\
        Ionosphere             & 351              & 33            & 126 (35.9\%)         \\
        Mammography            & 11183            & 6             & 260 (2.3\%)           \\
        NSL-KDD                & 148517           & 122           & 77054 (51.8\%)       \\
        Optdigits              & 5216             & 64            & 150 (2.9\%)           \\
        Pendigits              & 6870             & 16            & 156 (2.3\%)           \\
        Pima                   & 768              & 8             & 268 (34.9\%)         \\
        Satellite              & 6435             & 36            & 2036 (31.7\%)        \\
        Satimage-2             & 5803             & 36            & 71 (1.2\%)            \\
        Shuttle                & 49097            & 9             & 3511 (7.1\%)         \\
        Thyroid                & 3772             & 6             & 93 (2.5\%)            \\
        Wbc                    & 278              & 30            & 21 (7.6\%)            \\
        Wine                   & 129              & 13            & 10 (7.8\%)            \\
        \toprule
    \end{tabular}}
\end{table}

\bibliography{ref}
\bibliographystyle{IEEEtran}


\end{document}